\documentclass[10pt,twocolumn,letterpaper]{article}

\usepackage{iccv}
\usepackage{times}
\usepackage{epsfig}
\usepackage{graphicx}
\usepackage{amsmath}
\usepackage{amssymb}

\usepackage[breaklinks=true,bookmarks=false]{hyperref}

\iccvfinalcopy 

\def\iccvPaperID{****} 

\ificcvfinal\fi

\begin{document}

\title{Deep Learning Driven Detection of Tsunami Related Internal Gravity Waves: 
\\ a path towards open-ocean natural hazards detection}

\author{Valentino Constantinou\\
Terran Orbital Corportation\\
Irvine, California\\
{\tt\small valentino.constantinou@terranorbital.com}
\and
Michela Ravanelli\\
Institut de Physique du Globe de Paris\\
Paris, France\\
{\tt\small ravanelli@ipgp.fr}
\and
Hamlin Liu\\
University of California - Los Angeles\\
Los Angeles, California\\
{\tt\small hamlin.liu@gmail.com}
\and
Jacob Bortnik\\
University of California - Los Angeles\\
Los Angeles, California\\
{\tt\small jbortnik@atmos.ucla.edu}
}

\maketitle
\ificcvfinal\thispagestyle{empty}\fi

\begin{abstract}
    Tsunamis can trigger internal gravity waves (IGWs) in the ionosphere, perturbing the Total Electron Content (TEC) - referred to as Traveling Ionospheric Disturbances (TIDs) that are detectable through the Global Navigation Satellite System (GNSS). The GNSS are constellations of satellites providing signals from Earth orbit - Europe’s Galileo, the United States’ Global Positioning System (GPS), Russia’s Global’naya Navigatsionnaya Sputnikovaya Sistema (GLONASS) and China’s BeiDou. The real-time detection of TIDs provides an approach for tsunami detection, enhancing early warning systems by providing open-ocean coverage in geographic areas not serviceable by buoy-based warning systems.
    Large volumes of the GNSS data is leveraged by deep learning, which effectively handles complex non-linear relationships across thousands of data streams. We describe a framework leveraging slant total electron content (sTEC) from the VARION (Variometric Approach for Real-Time Ionosphere Observation) algorithm by Gramian Angular Difference Fields (from Computer Vision) and Convolutional Neural Networks (CNNs) to detect TIDs in near-real-time. Historical data from the 2010 Maule, 2011 Tohoku and the 2012 Haida-Gwaii earthquakes and tsunamis are used in model training, and the later-occurring 2015 Illapel earthquake and tsunami in Chile for out-of-sample model validation. Using the experimental framework described in the paper, we achieved a 91.7\% F1 score. Source code is available at: \textit{https://github.com/vc1492a/tidd}. Our work represents a new frontier in detecting tsunami-driven IGWs in open-ocean, dramatically improving the potential for natural hazards detection for coastal communities.
   
\end{abstract}

\section{Introduction}


It is widely acknowledged that natural hazards like earthquakes and tsunamis can produce acoustic and gravity waves able to propagate to the ionosphere \cite{Occhipinti2015, meng2019upper, astafyeva2019ionospheric, manta2020rapid}.
Tsunamis can trigger internal gravity waves (IGWs) which are amplified by the decreasing of the atmospheric density and can thus reach ionospheric heights, perturbing the electron content \cite{Daniels1952, Hines1972, Hines1974, Peltier1976}. 
These perturbations - Traveling Ionospheric Disturbances (TIDs) \cite{azeem2017traveling} - can be remotely detected by Global Navigation Satellite System (GNSS) derived measurements of the ionospheric Total Electron Content (TEC) \cite{giorgio, artru2005ionospheric}. 
Thus, ionospheric GNSS-TEC information can enhance tsunami warning systems \cite{ravanelli2021gnss, manta2020rapid, kamogawa2016possible}, mitigating disaster response issues by providing timely alerts and enabling prompt evacuation measures.
Indeed, GNSS-TEC can offer continuous, global updates on tsunami potential and arrival times reducing the risk of false alarms \cite{labrecque2019global}. Existing systems such as the Deep-ocean Assessment and Reporting of Tsunamis (DART) can be effective, but are limited to specific geographic locations due to the use of specialized hardware. This aim was expressed in the 2015 International Union of Geodesy and Geophysics resolution 4 (Real-Time GNSS Augmentation of the Tsunami Early Warning System) \cite{resolution}. 
The volume, variety and velocity of GNSS-TEC data provide a basis for machine and deep learning. The existence of several GNSS systems  brings ionospheric coverage that is increasing day by day, opening new perspectives for GNSS Ionospheric Seismology.

TID detection is a distinct multivariate time-series anomaly detection problem. Recently, Random Forest models have been explored to train models for detecting TIDs in TEC data, requiring feature engineering leveraging of both TEC and ionospheric spectrograms \cite{Brissaud2021}. Lately, deep learning has automatically discovered complex, highly non-linear features without requiring domain knowledge \cite{Munir2019}, resulting in greater modeling capability and flexibility versus other modeling approaches.

We leverage Gramian Angular Difference Fields (GADFs) and Convolutional Neural Networks (CNNs) for TID detection. The VARION (Variometric Approach for Real-Time Ionosphere Observation) algorithm is used to analyse the TEC time series \cite{ravanelli2021gnss} from multiple tsunamigenic earthquake events. We describe a framework for using deep learning to detect TIDs and assess the generalizability of the framework by utilizing multiple, separate events for model training and future event for validation. The 2010 Maule, the 2011 T\={o}hoku, the 2012 Haida-Gwaii earthquakes were used for model training with the 2015 Illapel earthquake used for out-of-sample validation.

\begin{figure} 
\centering
\includegraphics[height=1.8in]{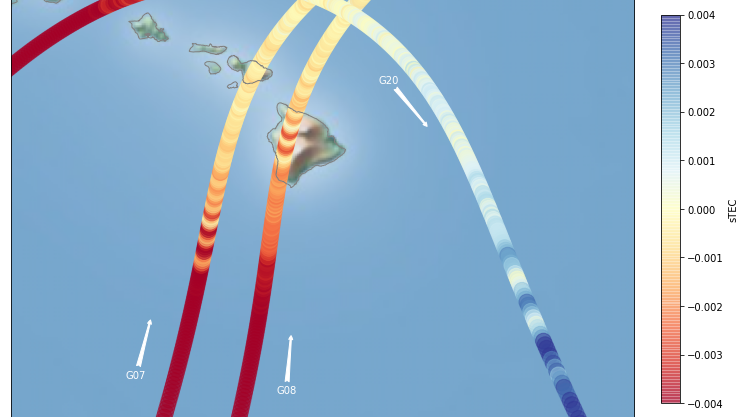}
\label{geoplot_gopm}
\caption{Map representing $\delta sTEC/\delta t$ [TECU/s] variations at the sub-Ionospheric Pierce Points for the 2012 Haida-Gwaii tsunami near the Hawaiian islands. 
The values recorded from the \textit{gopm} ground station for G07, G08 and G20 are shown.
}
\end{figure}

\section{Earthquake Information and Dataset}

The 2010 Mw. 8.8 Maule earthquake (36.1221$^\circ$S 72.8981$^\circ$W) \cite{usgs_maule} triggered a tsunami all over the Pacific region, reaching peaks of 29m at Constitución, Chile \cite{yue2014localized}. 
We employed a dataset of 34 days, 20 of which before the EQ and 10 after the EQ related to 30 GPS stations located in the Chile from the UNAVCO network \cite{unavco}. The shock triggered a tsunami that propagated all over the Pacific region, reaching over 700 kilometers of coastline \cite{ingv_maule}. The maximum run-up (i.e., the maximum topographic height reached by the tsunami) peak (29 meters) was recorded at Constitución, Chile \cite{yue2014localized}. The Paciﬁc Tsunami Warning Center (PTWC) issued the warning 12 minutes after the EQ \cite{soule2014post}. The tsunami arrived within 30 minutes at many locations in Chile, therefore, official evacuations and warnings by local authorities were not available at many places prior to the arrival of the tsunami \cite{itic}. The tsunami accounts for 124 victims concentrated in the coastal regions of Maule and Biobío, Juan Fernández Archipelago’s Robinson Crusoe Island and Mocha Island \cite{fritz2011field}.\\
The 2011 Mw 9.1 Great T\={o}hoku-Oki earthquake (38.297$^\circ$N 142.373$^\circ$E) \cite{usgs_tohoku}  generated tsunami 
waves (maximum of 20m) reaching the Pacific coast of Honshu within about 20 minutes and was observed all over the Pacific region \cite{itic_tohoku}. 1200 GNSS stations belonging to the GEONET network \cite{GEONET} were used to analyse
30 days, 10 of which before the EQ and 19 after the EQ. The shock triggered powerful tsunami waves that struck the Pacific coast of Honshu within about 20 minutes and that was observed all over the Pacific region. 
15270 and and 8499 people were reported to be killed and missed respectively because of the earthquake and tsunami.
In Sendai, maximum tsunami run-up heights (15-20 m range) were registered.
The Japan Meteorological Agency’s national tsunami warning center issued a tsunami warning 3 minutes after the earthquake triggering the alerting process that immediately broadcasted by mass media and locally activated sirens and other mitigation countermeasures such as flood gate closures. Nevertheless, many casualties resulted: waves overtopped tsunami walls and destroyed many structures, especially wooden homes \cite{itic_tohoku}.\\
The 2012 Mw 7.8 Haida Gwaii (52.788$^\circ$N 132.101$^\circ$W) earthquake \cite{usgs_haida} engendered a non-destructive tsunami registered throughout the Pacific. Tsunami waves up to 1.5m were registered in Maui and the Hawaii Island \cite{itic_haida}. 56 GPS stations placed on Hawaii islands belonging to UNAVCO network were used to study 15 days: 12 days before the EQ and 2 after the EQ. The quake engendered a non-destructive tsunami that was registered throughout the Pacific, hitting the coast of Alaska, of  British Columbia, of California and of Hawaii. 
The PTWC issued a tsunami warning (19:09 HST 27 October) that was then downgraded (01:01 HST 28 October).\\
The 2015 Mw 8.3 Illapel earthquake (31.573$^\circ$S 71.674$^\circ$W) \cite{usgs_illapel} provoked tsunami waves up to 9m. 80 GPS stations located in Chile from CSN \cite{csn} were used to investigate 26 days starting from 18 days before the EQ to 6 days after.  The tremor generated a tsunami that spread across the Pacific Ocean.
Tsunami waves heights up to 9 m on the coast were measured between 29$^\circ$S and 32$^\circ$S and smaller further south and north. Along the Chilean coast, the PTWC and National Hydrographic and Oceanic Service (SHOA) issued tsunami threat messages 7 and 8 min following the earthquake, respectively. Tsunami linked casualties were minimized by these prompt messages and evacuation \cite{satake2017review}.

\begin{table}[htb]
\centering
\begin{tabular}{lcc}
\multicolumn{3}{c}{\textbf{Earthquake Characteristics}}                   \\ \hline
                  & \textbf{year}    & \textbf{magnitude}    \\
\textit{Maule} & 2010                 & 8.8                 \\
\textit{Tohoku} & 2011                 & 9.1                 \\
\textit{Haida-Gwaii} & 2012                 & 7.8                 \\
\textit{Illapel}    & 2015               & 8.3                \\ \hline
                  & \multicolumn{1}{l}{} & \multicolumn{1}{l}{} 
\end{tabular}
\label{earthquake_characteristics}
\caption{The years and magnitudes of earthquakes in the training (2010 - 2012) and validation data (2015).}
\end{table}

\section{Methods}

\textbf{TEC Data, Labeling and Transformation:} We used VARION-produced \cite{ravanelli2021gnss} slant Total Electron Content variations over time ($\delta sTEC/\delta T$). VARION is based on single time differences of geometry-free combination of GNSS carrier-phase measurements, using a standalone GNSS receiver and standard GNSS broadcast products that are available in real time. VARION has been successfully applied to detection of ionospheric perturbations in several real-time scenarios \cite{ravanelli2020tids, ravanelli_2023}. Tsunami-TID periods range from 10 to 30 minutes \cite{astafyeva2019ionospheric}. TIDs generated from tsunami waves are similar to ionospheric disturbances generated from other phenomena, such as meteorite explosions in the atmosphere \cite{Yang2014, Luo2020}, 
volcanic eruptions \cite{Igarashi1994, Rozhnoi2014} 
or large explosions \cite{Kundu2021}.

The GNSS data stream source data exists at 5 second time intervals, but data was resampled to the minute level for modeling and demonstrating this proof of concept. Resampling the data reduces the computational resources needed to train machine and deep learning models. However, care must be taken when resampling as to not dampen behaviors in the time series important for detecting disturbances in time series. 

\begin{figure} 
\centering
\includegraphics[height=1in]{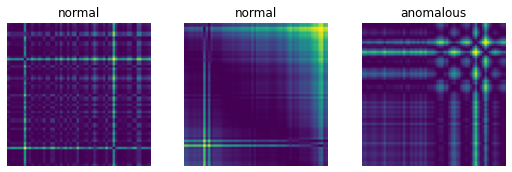}
\caption{A randomly-sampled batch of GADF-generated images from both the \textit{normal} and \textit{anomalous} classes.}
\label{batch_input_training}
\end{figure}

Supervised models were used, necessitating subject matter expert (SME)
ground-truth labels. Each univariate TEC time series was labeled with a start and finish time of the TIDs. For each time series $X = \{\textbf{x}^{(1)}, \textbf{x}^{(2)}, \ldots, \textbf{x}^{(n)}\}$ of $n$ TEC estimates, the data was split into windows of size $w$. Each window of TEC estimates are converted to images using GADF \cite{Wang2015}. GADF produces visually-interpretable differences across classes (Figure \ref{batch_input_training}). This data was then used in model training and validation. If \textit{any} of the image was generated from a window of data overlapping ground truth ranges, this image is categorized as representing a TID - if not, normal ionospheric TEC.

\subsection{Convolutional Neural Networks (CNNs)}

In this framework, data from multiple univariate time series are converted into images and utilized to train a deep learning model. Since information and characteristics about the time series are transformed into images in this framework, a computer vision approach is leveraged - a Convolutional Neural Network (CNN). A variety of deep learning architectures have been developed for image data, including CNNs. These computer vision models such as CNNs have been extensively used for natural hazards detection and disaster management \cite{Nijhawan2018, Said2019, Arshad2019, Rashidian2019}, from leveraging satellite optical imagery to multispectral data and even data from Synthetic Apertrure Radar (SAR). In this work, a single CNN model is trained for TID detection across data from the first three tsunami events in our data using the ResNet \cite{He2015} architecture, providing the model with the exposure to many different scenarios. 
This single model is then validated with unseen data from the 2015 Illapel earthquake and tsunami.

\subsection{A False Positive Mitigation (FPM) Strategy}\label{sec:false_positive_mitigation}

Time periods predicted as TIDs may be short. They may represent false positives or periods classified as TIDs, but are often representative of noise or normal behavior. . 
Our strategy considers the set of all univariate time series for a satellite $\textbf{X\textsubscript{s}}$, which contains many time series $X\textsubscript{g}$ of $n$ TEC estimates, one time series $X$ for each ground station. A boolean vector $\textbf{X\textsubscript{s,t}}$ of $1 x g$ dimension, where $g$ is the number of ground stations, is generated by representing TID detections as 1s and normal behavior as 0s. Vector values are summed and divided by $g$, generating a float value $F\textsubscript{s,t}$ between [0,1] representing the share of data at each time step representing a possible TID. A threshold $T\textsubscript{s,t}$ is selected such that any time periods $t$ where $F\textsubscript{s,t} > T\textsubscript{s,t}$ are considered as TIDs. 
The threshold parameter $T\textsubscript{s,t}$ is adjustable, with higher values reducing recall but improving precision. This approach ensures some level of agreement is reached across ground stations for a TID detection. 

\section{Results}

Training and validation metrics are reported, along with dataset summary statistics. We emphasized evaluating approach effectiveness.
No comparisons are made between CNN architectures, such as DenseNet or VGG.
Similarly, no comparisons are made between various types of image encoding methodologies. A more balanced dataset was created by undersampling the normal (majority) class such that the minority class represented 10\% of the number in the normal class (Table \ref{class_sample_sizes}). Increasing the overall share of the minority class relative to the size of the dataset is a commonly used technique for training generalized machine and deep learning models \cite{Yao2022}.

\begin{table}[htb]
\centering
\begin{tabular}{lcc}
\multicolumn{3}{c}{\textbf{Number of Samples}}                   \\ \hline
                  & \textbf{original}    & \textbf{balanced}    \\
\textit{anomalous} & 90201                 & 90201                 \\
\textit{normal}    & 18756848               & 900329                \\ \hline \\
\end{tabular}
\caption{The number of samples available in each class in the original data and the balanced dataset.}
\label{class_sample_sizes}
\end{table}

\subsection{Setup}

Being aware that ionospheric conditions are variable and earthquake features are different, we used 2 sets of tsunamigenic earthquakes. The dataset is used in model training, using data available from the 2010 Maule, 2011 T\={o}hoku and 2012 Haida-Gwaii earthquakes. Training data is randomly sampled for training and testing sets used for model training, with precision, recall, and F1-score metrics provided (Table \ref{metrics}) from the training process. 

\begin{table}[htb]
\centering
\begin{tabular}{lc}
\multicolumn{2}{c}{\textbf{Model Parameters}}    \\ \hline \\[-1em] 
architecture               & resnet50            \\
batch size                 & 512                  \\
optimizer                  & Adam                \\
beginning learning rate    & 0.00025            \\ 
loss function              & Cross-Entropy Loss \\
image size (in pixels)  & 224x224 \\ \hline \\
\end{tabular}
\caption{The parameters of the model used in the experiments.}
\label{paremeters}
\end{table}

The second dataset contains tsunami-induced TIDs occurring after the events in the training set, simulating how such a model would be used in the real-world. Each minute, data streams are processed chronologically with 60-minute windows of sTEC data, with a GADF-generated image predicted by the trained model to contain a TID or not. Labels are concatenated, producing labeled sequences. Each ground truth anomalous sequence $x_{a} \in \textbf{x}_{a}$ of values is then  evaluated against the set of predicted sequences, according to the rules described in \cite{Hundman2018}.
 The FPM strategy is leveraged compared to the standard approach in Table \ref{metrics}.

\subsection{Model Parameters and Evaluation}

The CNN architecture and parameters are described in Table \ref{paremeters}. 
A sequence length of $l_s = 60$ minutes is used. Steps are taken to minimize over-fitting, such as reducing the learning rate on loss plateauing and using early stopping. 
The threshold $T\textsubscript{s,t}$ used in FPM was kept constant at 0.75.

\section{Discussion and Conclusion}

\begin{table}[htb]
\centering
\begin{tabular}{lc}
\multicolumn{2}{c}{\textbf{Metrics}}    \\ \hline \\[-1em] 
\textit{validation} - recall        & 96.2.\%                \\
\textit{validation} - precision     & 34.7\%                \\
\textit{validation} - F\textsubscript{1} score      & 51.0\%     \\

\textit{validation (false positive mitigation)} - recall        & 84.6.\%        \\
\textit{validation (false positive mitigation)} - precision     & 100.0\%                \\
\textit{validation (false positive mitigation)} - F\textsubscript{1} score      & \textbf{91.7\%}     \\

\hline \\
\end{tabular}
\caption{Performance metrics from various stages of the experiment.}
\label{metrics}
\end{table}

\begin{figure}
\includegraphics[width=3.15in]{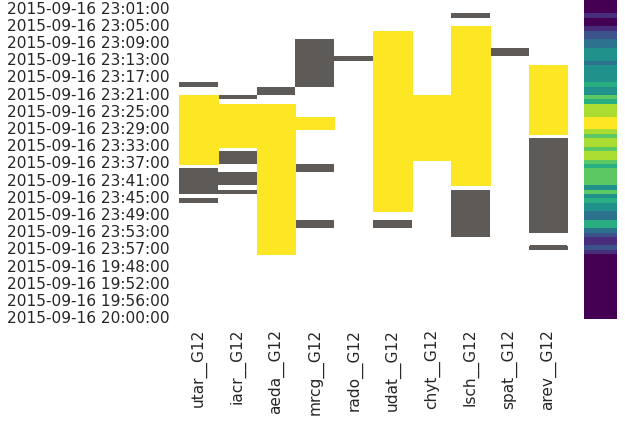}
\caption{TID classifications for satellite G12 and its ground stations. Lighter-shaded regiions are considered as TIDs following FPM, with those in the darker shade no longer considered as TIDs. The weighted score $F\textsubscript{s,t}$ is shown on the right.}
\label{false_positive_mitigation_G12}
\end{figure}

\textbf{Impacts of the FPM Strategy:} The FPM strategy has a profound impact on reducing false positives, improving the precision from 34.7\% to 100.0\% and F\textsubscript{1} score from 51.0\% to 91.7\% (Table \ref{metrics}). The 91.7\% F\textsubscript{1} score is achieved with a smaller amount of events compared to earlier work \cite{Brissaud2021}. Darker shades in Figure \ref{false_positive_mitigation_G12} show TID classifications that are re-classified to normal behavior. Lighter shades indicate time periods considered a TID. 
The ideal threshold $T\textsubscript{s,t}$ can be selected based on requirements. For example, a cautious system may provide false positives and capture all potential instances of a TID. However, a system tuned this way is not a practical alerting system. In general, a higher threshold reduces the false positives (improving precision) but decreases recall.

\textbf{Good Data Delivers:} A single set of labels was produced for each event by a SME. This expert utilized her scientific knowledge to represent TID start and finish times. Utilizing labels from multiple SMEs - together with a human-in-the-loop (HIL) process - would provide a refined perspective on what constitutes a TID. Broad consensus is better than a single opinion, and this is best captured in data by utilizing multiple subject matter experts for labeling. 

Additional performance gains are achievable using from more events and geographic areas. Data from other events could be considered.
While parameters such as the model parameters, batch size or FPM strategy are adjustable, large gains in model performance are best achieved by leveraging a larger number of historical events. Future work should consider continuing to focus on data preparation and management, from increasing the number of events used for training and validation, to improving data labeling processes - important considerations for a real-world system.  

\textbf{Detecting Other TIDs:} It is important to note that future deep learning analyses should also focus on the identification of the different kinds of ionospheric perturbations. 
Indeed, while this framework was developed with the intent of detecting tsunami-induced TIDs, this approach can be used more broadly to detect ionospheric perturbation from different sources (creating a multi-classification problem) or time series anomalies in other domains.\\ 
Finally, the joint application of deep learning and GNSS-TEC observations can effectively contribute to the enhancement of tsunami early warning systems and hence to improve disaster response procedures.

{\small
\bibliographystyle{egpaper_final}
\bibliography{egpaper_final}
}

\end{document}